\documentclass[conference]{IEEEtran}
\IEEEoverridecommandlockouts

\usepackage{xspace}
\usepackage{tcolorbox}
\usepackage{enumitem}
\usepackage[ruled]{algorithm2e}

\usepackage[numbers,sort]{natbib}
\usepackage[colorlinks=true,citecolor=blue,linkcolor=blue,urlcolor=blue,bookmarks=false]{hyperref}

\usepackage{booktabs} 
\usepackage{multirow}
\usepackage{graphicx}
\usepackage{tabularx}
\newcolumntype{C}[1]{>{\centering\arraybackslash}p{#1}}
\usepackage{colortbl}
\usepackage{graphicx}  
\usepackage{float}  
\usepackage{subfig}
\usepackage{makecell} 
\usepackage{color}
\usepackage{arydshln} 
\usepackage{caption}
\usepackage{wrapfig}
\usepackage{colortbl}
\usepackage{amsthm} 
\usepackage{amsmath}
\usepackage{adjustbox}
\usepackage{threeparttable}
\usepackage{stfloats}
\usepackage{pifont}
\usepackage{xurl}
\usepackage{amsfonts}

\theoremstyle{definition}

\usepackage{microtype}
\newcommand{\circnum}[1]{\raisebox{.5pt}{\textcircled{\raisebox{-.25pt}{\scalebox{0.8}{#1}}}}}
\newcommand{\ie}[0]{\textit{i.e.,}\xspace}
\newcommand{\eg}[0]{\textit{e.g.,}\xspace}

\newcommand{\tool}{\textsc{Argus}\xspace}
\newcommand{\tools}{\textsc{Argus}$^*$\xspace}

\newcommand{\TCPtool}{TCP-\textsc{Argus}$^*$\xspace}
\newcommand{\UniADtool}{UniAD-\textsc{Argus}$^*$\xspace}
\newcommand{\VADtool}{VAD-\textsc{Argus}$^*$\xspace}

\newcommand{\TCPtools}{TCP-\textsc{Argus}\xspace}
\newcommand{\UniADtools}{UniAD-\textsc{Argus}\xspace}
\newcommand{\VADtools}{VAD-\textsc{Argus}\xspace}
\newcommand{\Apollotools}{Apollo-\textsc{Argus}\xspace}

\newcommand{\todo}[1]{\textcolor{black}{#1}}

\newcommand{\mr}[1]{\textcolor{black}{#1}}
\def\BibTeX{{\rm B\kern-.05em{\sc i\kern-.025em b}\kern-.08em
    T\kern-.1667em\lower.7ex\hbox{E}\kern-.125emX}}
\begin{document}
\title{\tool: Resilience-Oriented Safety Assurance Framework for End-to-End ADSs}


\author{
\IEEEauthorblockN{
  Dingji Wang\IEEEauthorrefmark{1}, You Lu\IEEEauthorrefmark{1}, Bihuan Chen\IEEEauthorrefmark{1}\IEEEauthorrefmark{2}, 
  Shuo Hao\IEEEauthorrefmark{1}, Haowen Jiang\IEEEauthorrefmark{1}, 
  Yifan Tian\IEEEauthorrefmark{1}, 
  Xin Peng\IEEEauthorrefmark{1}
}
\IEEEauthorblockA{\IEEEauthorrefmark{1}College of Computer Science and Artificial Intelligence, Fudan University, Shanghai, China}
\IEEEauthorblockA{\IEEEauthorrefmark{2}Corresponding Author}
}

\maketitle

\begin{abstract}
End-to-end autonomous driving systems (ADSs),~with their strong capabilities in environmental perception and generalizable driving decisions, are attracting growing attention from both academia and industry. However, once deployed on public roads, ADSs are inevitably exposed to diverse driving hazards that may compromise safety and degrade system performance. This raises a strong demand for resilience of ADSs, particularly the capability to continuously monitor driving hazards and adaptively respond to potential safety violations, which is crucial for maintaining robust driving behaviors in complex driving scenarios.

To bridge this gap, we propose a resilience-oriented runtime framework, named \tool, to mitigate the driving hazards,~thus preventing potential safety violations and improving the driving performance of an ADS. \tool continuously monitors the trajectories generated by the ADS for potential hazards and, whenever the EGO vehicle is deemed unsafe, seamlessly takes control~via a hazard mitigator.
We integrate \tool with~three state-of-the-art end-to-end ADSs, \ie TCP, UniAD and VAD. Our evaluation has demonstrated that \tool effectively and efficiently enhances the resilience of ADSs, improving the driving score of ADSs by 150.30\% on average, and preventing 64.38\% of the violations, with little additional time overhead.
\end{abstract}



\section{Introduction}\label{intro}

Recently, the superior environmental perception and generalizable decision-making capabilities of end-to-end autonomous driving systems (ADSs), enabled by specialized model architectures and large-scale diverse training data~\cite{hu2023planning, jiang2023vad, shao2023safety}, have attracted substantial investment from both academia and industry~\cite{wu2022trajectoryguided, hu2023planning, jiang2023vad}. Unlike traditional modular pipelines~\cite{apollo}, end-to-end ADSs do not rely on high-definition maps or extensive hand-crafted rules to handle diverse driving scenarios. 
These ADSs promise to enhance road safety, mitigate traffic congestion,~and boost overall transportation efficiency, thereby catalyzing~transformative change across the automotive industry~\cite{paden2016survey}.
\mr{However, they lack explicit logic explainability~and~safety boundaries compared to modular ADSs.}
Scenario-based testing approaches~\cite{cheng2023behavexplor, zhang2023building, lu2024advfuzz, lu2024diavio} demonstrate that the ADSs may still suffer from safety violations even in scenarios that are highly similar to those in the training dataset, compromising safety and degrading system performance, as highlighted~by~numerous documented incidents~\cite{tesla, xiaomi}. 
Therefore, it is important to enhance the resilience of ADSs, achieving the hazard mitigation while maintaining their advantages in~interactive~scenarios.

\mr{Software resilience focuses on the capability of systems to adapt to and recover from unexpected events while maintaining effective operation under hazardous conditions~\cite{hollnagel2006resilience, provan2020safety}.} Improving the resilience requires not only monitoring hazards but also building systems that can proactively mitigate hazards, recover quickly, and continue to operate effectively in complex conditions~\cite{righi2015systematic}. Several studies have explored approaches to enhance resilience across diverse domains, including aviation, robotics,~and~connected autonomous vehicle platoons~\cite{camara2015robustness, wan2022analyzing, fang2024resilience}.

As to end-to-end ADSs, only a few works~\cite{bogdoll2022anomaly, chia2022risk, grewal2024predicting}~focus on unexpected condition prediction~\cite{stocco2022thirdeye, stocco2020misbehaviour} or rule-based misbehavior identification of ADSs~\cite{candela2023risk,candela2021fast,qian2025collision,yu2024online}, but fail~to provide a runtime solution for hazard mitigation. To prevent safety violations, several works~\cite{sun2024redriver,sun2025fixdrive} make intrusive modifications to original ADSs and refine the trajectories generated~by ADSs to ensure the specification compliance. Furthermore, various fallback strategies~\cite{yu2019fallback, hussain2022deepguard,xue2018fallback}, \ie~emergency braking~\cite{kusano2012safety, zhang2024dual} and human-initiated takeovers~\cite{gold2013take,yu2022remote}, have been proposed to ensure safety. These approaches primarily emphasize emergency interventions rather than enabling ADSs to automatically adapt and maintain effective operation during hazardous scenarios. However, according to the SAE levels~\cite{SAE} of automation, the fallback responsibility no longer lies with human drivers but instead must be handled by the ADS itself for Level 4+~\cite{yu2019fallback}. To the best of our knowledge, there is no existing work that explicitly targets enhancing the resilience of end-to-end ADSs, particularly the capability to continuously monitor hazards, adaptively respond to potential safety violations, and recover quickly to sustain safe operation in hazardous scenarios.

To fill this gap, we present \tool, a  resilience-oriented runtime framework that proactively prevents safety violations and enhances the driving performance of an end-to-end ADS. \tool comprises three components, \ie the~\textit{Takeover~Gate},~the \textit{Hazard Monitor}, and the \textit{Hazard Mitigator}. Every trajectory generated by the ADS passes through the \textit{Takeover Gate}, which is responsible for dynamic control switching between the ADS and the \textit{Hazard Mitigator} that is built upon the intelligent driver model (IDM)~\cite{treiber2000congested}.
The \textit{Takeover Gate} leverages the takeover and recovery buffers maintained by the \textit{Hazard Monitor} to determine whether a given trajectory is safe for execution, thereby realizing a hazard-aware takeover and recovery mechanism.
If the EGO vehicle (\ie the vehicle controlled by the ADS) is deemed unsafe, control is taken over by the \textit{Hazard Mitigator} for hazard mitigation, and is only returned to the ADS once safety is reestablished.

We have conducted large-scale experiments to evaluate the effectiveness and efficiency of \tool. First, we integrate \tool with three state-of-the-art end-to-end ADSs (\ie TCP~\cite{wu2022trajectoryguided}, UniAD~\cite{hu2023planning} and VAD~\cite{jiang2023vad}), and evaluate them on two benchmarks (\ie Bench2Drive~\cite{jia2024bench} and CARLA leaderboard 2.0 validation set~\cite{CARLA2.0}) with realistic perception-based BEVs and the ideal privileged environment information, respectively. Our experiments have demonstrated that \tool enhances the resilience of ADSs, improving the driving score of ADSs by 150.30\% on average, with little additional time overhead. Besides, \tool~is able to prevent 64.38\% of the violations with better driving skills.
Second, we determine the accuracy of these takeovers triggered by \tool, and the results has shown that \tool achieves a high $\text{F}_3$ score of \todo{0.899} in producing takeover decisions. Then, we conduct ablation studies to assess the individual contribution of each step within the IDM-based hazard mitigator, \mr{followed by the parameter sensitivity analysis to evaluate the robustness of \tool in different parameter settings}. \mr{Finally, we generalize \tool to Apollo~\cite{apollo}, a modular ADS, where \tool achieves a 113.92\% improvements over the original Apollo.}

\mr{Overall, \tool introduces a modular, resilience-oriented safety assurance framework for AI-enabled complex systems with probabilistic uncertainty, embodying key software engineering principles (\ie runtime monitoring and adaptive recovery).}
The main contributions of our work are as follows.
\begin{itemize}[leftmargin=*]
    \item We propose a hazard-aware takeover and recovery mechanism and an IDM-based hazard mitigator to prevent safety violations and enhance the performance of end-to-end ADSs.
    \item We design and implement a framework \tool to enhance the resilience of end-to-end ADSs under driving hazards.
    \item We integrate \tool with three state-of-the-art end-to-end ADSs, and conduct experiments on two benchmarks to demonstrate the effectiveness and efficiency of \tool.
\end{itemize}

\section{Background}

\subsection{End-to-End Autonomous Driving Systems}
End-to-end ADSs~\cite{chen2024end, sun2023interpretable} unify perception, prediction, and planning within a single model. These models process multi-modal sensor inputs (\eg multi-view camera images and radar), navigation points, and vehicle states (\eg speed, heading, and position) to plan future trajectories composed of short-term waypoints and the desired speed.~A vehicle controller, typically a PID-based controller~\cite{borase2021review}, subsequently converts every trajectory generated by the model into control commands (\ie brake, steer and throttle) to drive the vehicle. 

Beyond performance improvement achieved by simply scaling training resources, recent work~has proposed in-model safety mechanisms to enhance the robustness of end-to-end ADSs. For instance, Interfuser~\cite{shao2023safety} formulates a linear program to optimize expected speed for collision avoidance. UniAD~\cite{hu2023planning} employs Newton-based optimization over occupancy grids to generate collision-free trajectories. In addition, VAD~\cite{jiang2023vad} introduces instance-level planning constraints to improve planning reliability. However, recent studies~\cite{jia2024bench, sima2025drivelm} have shown that end-to-end ADSs still lack resilience, with violations occurring even in scenarios encountered during training. 

\subsection{Intelligent Driver Model}\label{sec:IDM}

The intelligent driver model (IDM)~\cite{treiber2000congested} is a widely used car-following model that computes positions and speeds based on a leading actor in front, typically serving as a component within rule-based driving models~\cite{albrecht2021interpretable,krajzewicz2002sumo}. 
For the EGO vehicle $E$ and the selected leading actor $lead$, $p_E$ denotes the position of $E$ at time $t$, and $v_E$ denotes its speed. Furthermore, $s_E$ is defined as the net distance between $E$ and $lead$, and $\Delta v_{E}$ represents their relative speed.
For a simplified version of the model, the dynamics of $E$ are described by Eq.~\ref{IDM},
\begin{equation}\label{IDM}
\begin{array}{ll}
    \dot{p}_{E}=\frac{\mathrm{d} p_{E}}{\mathrm{d} t}=v_{E}\\
    \dot{v}_{E}=\frac{\mathrm{d} v_{E}}{\mathrm{d} t}=a\left(1-\left(\frac{v_{E}}{v_{0}}\right)^{\sigma}-\left(\frac{s^{*}\left(v_{E}, \Delta v_{E}\right)}{s_{E}}\right)^{2}\right) \\
    \text{with }s^{*}\left(v_{E}, \Delta v_{E}\right)=s_{0}+v_{E} T+\frac{v_{E} \Delta v_{E}}{2 \sqrt{a b}}
\end{array}
\end{equation}
where $v_0$ denotes the desired speed, $s_0$ is the minimum desired net distance, $T$ is the expected minimum time required for $E$ to reach the position of $lead$, $a$ is the maximum acceleration, $b$ is the comfortable braking deceleration, and $\sigma$ is the acceleration exponent that is usually set to 4.

Although the IDM has been extensively validated and is known for its effectiveness in modeling realistic traffic flow and longitudinal driving behavior, it remains limited in its applicability to hazard mitigation. In particular, its reliance on a single fixed leading actor and the absence of dynamic route adaptation limit its capability to navigate~diverse~driving~hazards.

\section{Methodology}

\begin{figure*}[!t]
    \centering
    \includegraphics[width=0.85\linewidth]{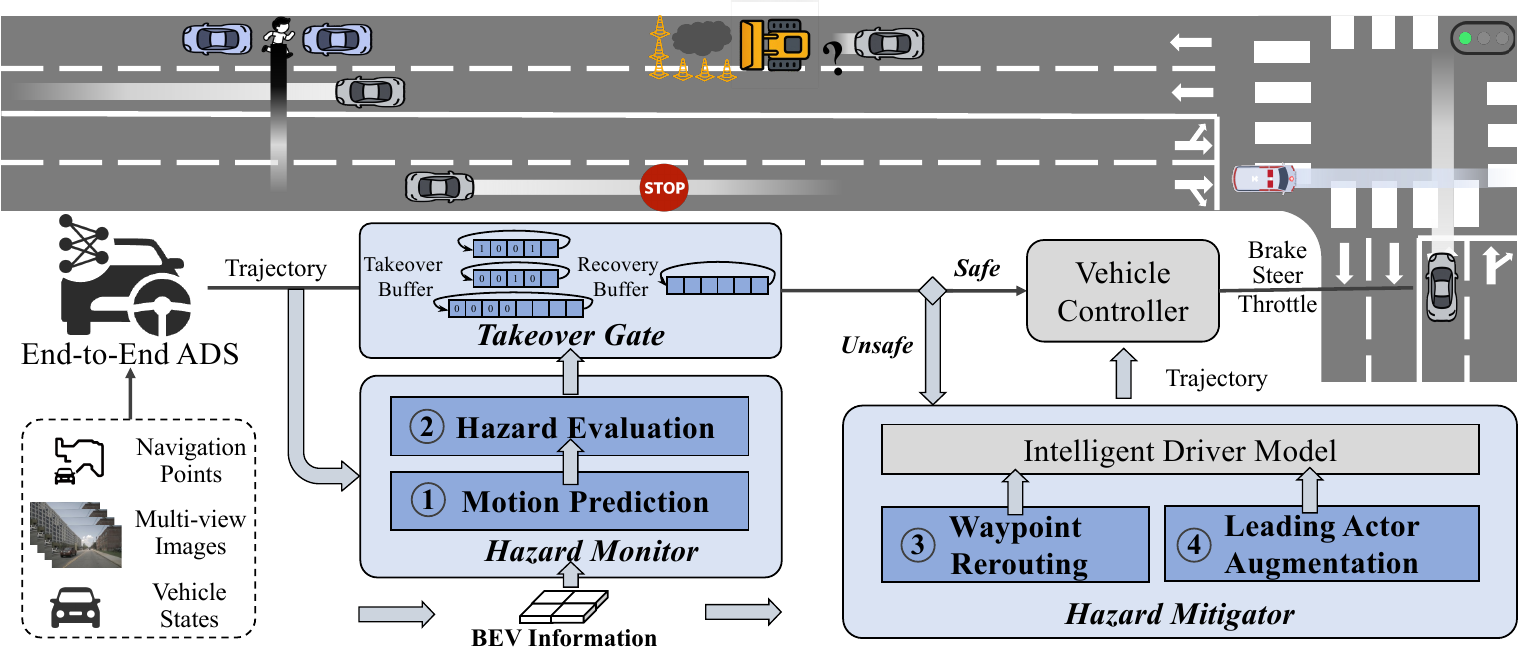}
    \caption{Approach Overview of \tool}
    \label{img:overview}
\end{figure*}

We propose \tool to enhance the resilience of an end-to-end ADS under hazardous driving scenarios. Our framework targets three primary hazards, \ie the \textit{collision hazard} where the EGO vehicle collides with traffic participants, the \textit{stop signal hazard} where the EGO vehicle violates the stop signals (\eg stop signs and red lights), and the \textit{stalling hazard} where the EGO vehicle is stalled in a dangerous situation.
The overall idea of \tool is to asynchronously monitor the trajectories generated by the ADS for potential hazards, and to take over control using an IDM-based hazard mitigator when the EGO vehicle is deemed unsafe. Control is returned to the ADS only after the safety is~reestablished.

\subsection{Approach Overview}

Fig.~\ref{img:overview} shows the approach overview of \tool, which consists of three components, \ie the \textit{Takeover Gate}, the \textit{Hazard Monitor}, and the IDM-based \textit{Hazard Mitigator}. Every trajectory (\ie the desired waypoints and speed) generated by the ADS passes through the \textit{Takeover Gate} (see Sec.~\ref{Takeover Checker}),~which checks three takeover buffers and one recovery buffer maintained by the \textit{Hazard Monitor}, to determine whether it is safe for execution by a takeover and recovery mechanism. If the~EGO vehicle is unsafe, the \textit{Hazard Mitigator}  is activated~for~mitigation. 

The \textit{Hazard Monitor} is responsible for asynchronous hazard detection based on trajectories generated by the ADS (see Sec.~\ref{Hazard Monitor}). Specifically, it utilizes the Bird's Eye View~(BEV) representations derived from sensor data (\ie multi-view~camera images)~to predict the motions of the EGO vehicle itself~as well as~the surrounding traffic participants (\ie step \circnum{1} in Fig.~\ref{img:overview}). Subsequently, the \textit{Hazard Monitor} evaluates~the~potential hazards using the predicted motions and EGO vehicle states (\ie step \circnum{2} in Fig.~\ref{img:overview}), updating the values of the three takeover buffers and one recovery buffer for the \textit{Takeover Gate}.

The IDM-based \textit{Hazard Mitigator} is responsible for performing active hazard mitigation according to the current hazardous situation (see Sec.~\ref{Hazard Mitigator}). 
Specifically, it generates and optimizes the waypoints to avoid obstacles (\ie step \circnum{3} in Fig.~\ref{img:overview}), subsequently passing the optimized waypoints to the IDM. Then, the \textit{Hazard Mitigator} augments the leading actors along with the optimized waypoints (\ie step \circnum{4} in Fig.~\ref{img:overview}) for the IDM to compute the speed of the EGO vehicle and achieve dynamic route adaptation.
Finally, the resulting trajectory, composed of the optimal waypoints and speed, is sent to the vehicle controller for execution, thereby circumventing the potential~hazards.


\subsection{Takeover Gate}\label{Takeover Checker}
To determine whether the control of the EGO vehicle~should remain with the ADS or be temporarily taken over, we introduce the \textit{Takeover Gate}. It checks three takeover buffers and~one recovery buffer, deciding whether to dispatch the trajectory generated by the ADS to the vehicle controller. 

\textbf{Takeover Buffer.} Implemented as a fixed-length circular queue, the takeover buffer stores the most recent hazard evaluation results for ADS trajectories. It provides a running window for identifying whether the ADS is currently exposed to hazards. Two distinct queues with length $M$ are maintained for collision hazard and stop signal hazard,~respectively. Besides, a queue with length $N$ is maintained for stalling hazard. 

\textbf{Recovery Buffer.} Likewise, realized as a circular queue of length $R$, the recovery buffer records the latest $R$ hazard evaluations after a takeover. It tracks the ADS's progression from a hazardous state to a safe state, thereby determining when the control can be safely returned to the ADS.

\textbf{Takeover and Recovery Mechanism.} The takeover buffers and the recovery buffer are initially empty, and each of their entries is a binary value (\ie~0 or 1), denoting the~absence or presence of a potential hazard. The values in these buffers are continuously updated and maintained by the \textit{Hazard Monitor}, which will be introduced in Sec.~\ref{Hazard Monitor}. Leveraging the entries in these buffers, the \textit{Takeover Gate} determines whether a given trajectory should be executed by the vehicle controller. 

Specifically, if the number of entries with value 1 in either the collision hazard buffer or stop signal hazard buffer exceeds a threshold $l$, or all entries in the stalling hazard~buffer~are~1, the current trajectory is not dispatched to the vehicle controller, and the downstream \textit{Hazard Mitigator} is instead activated, which will be introduced in Sec.~\ref{Hazard Mitigator}.
The collision and~stop signal hazards are safety-critical, and thus a  threshold-based conservative takeover strategy is adopted to account for potential false negatives in \textit{Hazard Monitor}. The stalling hazard indicates that the EGO vehicle is persistently unable to proceed and therefore requires takeover.
Once the control has been taken over, the recovery buffer is reset and begins recording the evaluation results for subsequent ADS trajectories. When all entries in the recovery buffer are 0, indicating that the ADS has remained in a consistently safe state and~is~capable of resuming the driving task, the control is returned to the ADS by allowing its newly generated trajectory to be dispatched~to~the~vehicle~controller. 

To balance responsiveness and stability, we construct a~small calibration set from \cite{jia2024bench}, and empirically tune~the parameters $M$, $R$ and $l$ to adapt to different ADS~configurations.


\subsection{Hazard Monitor}\label{Hazard Monitor} \label{sec:HM}

To detect the potential driving hazards (\ie the collision hazard, the stop signal hazard, and the stalling hazard), for each trajectory generated by the ADS, we asynchronously assess whether the ADS falls under potential hazards. This process involves two key steps, \ie the \textbf{motion prediction}, which prepares the predicted future motions of the EGO vehicle and surrounding traffic participants, and the \textbf{hazard evaluation}, which evaluates the potential hazards and maintains the takeover buffers and the recovery buffer.

\begin{figure}[!t]
    \centering
    \includegraphics[width=.7\linewidth]{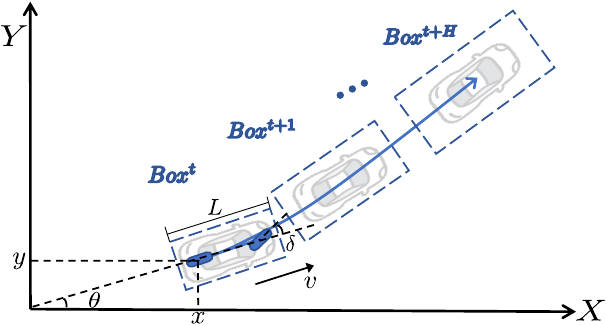}
    \caption{An Example of Motion Prediction for Vehicles}
    \label{img:KBM}
\end{figure}

\textbf{Motion Prediction.} 
As to vehicles, we adopt the kinematic bicycle model (KBM)~\cite{polack2017kinematic} to predict the future motions of the EGO vehicle itself and surrounding vehicles. As shown in Fig.~\ref{img:KBM}, KBM is a simplified model of vehicle dynamics, which models vehicles as two-wheel systems, capturing the essential characteristics of vehicle motions, including positions $p = (x, y)$, headings $\theta$, and speeds $v$. It employs nonlinear dynamics to iteratively compute the future motions of a vehicle over a period of frame in the future using Eq.~\ref{KBM}, 
\begin{equation}\label{KBM}
\frac{d}{d t}\left(\begin{array}{l}
x \\
y \\
\theta \\
v
\end{array}\right)=\left(\begin{array}{c}
v \cos (\theta) \\
v \sin (\theta) \\
v \tan (\delta) / L \\
a
\end{array}\right)
\end{equation}
where $a$, $\delta$ and $L$ are the acceleration, the steering angle, and the length of the vehicle, respectively. 

Specifically, we first obtain EGO vehicle states (\ie $p$, $v$ and $\theta$) from standard onboard sensors, and its physical dimensions (\ie vehicle length $L$ and width $W$) from vehicle specifications.
We parse the $a$ and $\delta$ of the EGO vehicle from the trajectory generated by the ADS at frame $t$. 
Besides, we utilize the BEV representation $BEV^t$ to get the states of the surrounding vehicles. The $BEV^{t}$ provides the $p$, $v$, $\theta$, $L$ and $W$ of each surrounding vehicles at frame $t$. We calculate the $a$ and $\delta$ of each surrounding vehicle using the difference between their speed and heading in $BEV^t$ and $BEV^{t-1}$, respectively. Then, we apply the KBM to predict the future motions of the EGO vehicle and surrounding vehicles over a frame duration $H$, which is set to 60 (\ie 3 seconds)~by default~\cite{sun2022lawbreaker}. Finally, we construct the bounding boxes $Box = (x, y, \theta, v, L, W)$ of the EGO vehicle and surrounding vehicles during $H$ by attaching their physical dimensions with predicted motions generated by the KBM. As shown in Fig.~\ref{img:KBM}, following~\cite{sima2025drivelm}, we linearly enlarge each predicted bounding box over frame, up to 130\% of the EGO vehicle's original dimensions and up to 200\% of surrounding vehicles, to mitigate accumulated~prediction~errors~of~KBM.

As to pedestrians, we model the movement of pedestrians with a constant speed assumption along current heading. We obtain the position, speed, heading and physical dimensions of each pedestrian from $BEV^t$ and construct their bounding boxes over the frame period $H$ similar to vehicles, enlarging the size of the original bounding box in $BEV^t$ by~50\%.

As to static obstacles, we directly extract the bounding boxes from the BEV representation $BEV^t$, which are assumed to remain stationary throughout the prediction horizon.

To monitor potential hazards of stop signal violation, inspired by Apollo~\cite{apollo}, the influence regions of stop signals (\eg stop signs and red lights) are abstracted as virtual static bounding boxes.
For instance, based on the EGO vehicle's heading,~as well as the position and orientation of the stop sign in $BEV^t$, a 3$\times$3 meter square region is constructed on the lane to represent the stop sign's influence region.~The~bounding~box~of~the~stop sign remains active until the EGO vehicle comes to a complete stop in this region, at which the bounding box is deactivated.

Finally, through motion prediction, we obtain a set of predicted bounding boxes of the EGO vehicle $E$ and all surrounding traffic participants $T$, including vehicles, pedestrians, static obstacles and stop signals, over the prediction~horizon $[t, t+H]$, denoted as $\mathbf{Box}^t = \{Box_\alpha^i \mid \alpha \in E \cup T$, $i \in [t, t+H]\}$, where each bounding box $Box_\alpha^i$ is denoted as $ Box_\alpha^i = (x_\alpha^i, y_\alpha^i, \theta_\alpha^i, v_\alpha^i, L_\alpha, W_\alpha)$.
All the predicted bounding boxes are used to evaluate the potential hazards in the~next~step.

\textbf{Hazard Evaluation.} As mentioned above, we focus on the evaluation of three primary hazards, \ie the \textit{collision hazard}, the \textit{stop signal hazard}, and the \textit{stalling hazard}.

\textit{Collision Hazard.} We detect potential collisions by assessing whether the EGO vehicle's predicted bounding boxes $Box_{E}^t$ result in unsafe or illegal interactions with other traffic participants. Specifically, we identify the intersections using the separating axis theorem~\cite{huynh2009separating}, which determines $\text{IS}(Box_i^t, Box_j^t) = \text{True}$ if and only if no separating axis exists between $Box_i^t$ and $Box_j^t$ at frame $t$. With respect to potential collisions, for the $\mathbf{Box}^t$, we calculate the minimum predicted collision frame $C^t$ of the EGO vehicle by Eq.~\ref{eq:tcoll}.
\begin{equation}\label{eq:tcoll}
\small
    C^t = \min \{i \in [t, t + H] \mid \exists \alpha \in T ,
    \text{IS}(Box_\text{E}^{i},\, Box_\alpha^{i}) = \text{True}\} 
\end{equation}
To reduce the false positive rate, we compare $C^t$ with $C^{t-1}$ that is calculated at previous frame. We consider a potential collision is expected to occur when $C^t \le C^{t-1}$, indicating that the potential collision is becoming more imminent and the ADS trajectory has not shown a tendency to avoid the hazard.

\textit{Stop Signal Hazard.} A stop signal violation is considered to occur when the EGO vehicle enters the influence region of a stop signal and fails to come to a complete stop (\ie its speed remains above a predefined threshold $\epsilon = 0.1$ throughout the entire period of intersection with the influence region). For the bounding boxes of the EGO vehicle $Box_{E}^i$, the speeds of the EGO vehicle $v_E^i$, and the bounding boxes of the stop signal $Box_{sg}^i$ over the prediction horizon $[t, t+H]$, we determine the $SigVio = \text{True}$ when Eq.~\ref{eq:signal} is satisfied.
\begin{equation}\label{eq:signal}
\small
\forall i' \in \{i \in [t, t + H]\mid \text{IS}(Box_{E}^{i},\, Box_{sg}^i) = \text{True}\} , v_E^{i'} > \epsilon
\end{equation}

\textit{Stalling Hazard.} A stalling hazard is considered to occur when the EGO vehicle slows to a near stop (\ie its speed~falls below the predefined threshold $\epsilon$) outside any influence region of stop signals at current frame $t$. Specifically, for the bounding box of the EGO vehicle $Box_{E}^t$ and its speed $v_E^t$ at frame $t$, we determine the $Stalling = \textit{True}$ when Eq.~\ref{eq:stall} is satisfied.
\begin{equation}\label{eq:stall}
    \small
    \begin{array}{cc}
     v_E^t < \epsilon \;\wedge\; \neg\, \text{ValidStop}(Box_{E}^t,t)
\end{array}
\end{equation}
where $\text{ValidStop}(Box_{E}^t, t)$ is a function that checks whether the EGO vehicle is in any influence region of stop signals derived from the BEV representation $BEV^t$ at frame $t$. 

Finally, we maintain the three takeover buffers in the \textit{Takeover Gate}, \ie the collision buffer $B_\text{collision}$ of size $M$, the stop signal buffer $B_\text{signal}$ of size $M$, and the stalling buffer $B_\text{stall}$ of size $N$, by Eq.~\ref{buffer},
\begin{equation}\label{buffer}
    \small
\begin{array}{ll}
B_{\text{collision}}[t \bmod M] = \mathbb{I}(C^t \le C^{t-1}) \\
B_{\text{signal}}[t \bmod M] = \mathbb{I}(SigVio = \text{True}) \\
B_{\text{stall}}[t \bmod N] = \mathbb{I}(Stalling = \text{True}) 
\end{array}
\end{equation}
where $\mathbb{I}(\cdot)$ is the indicator function, which returns 1 if the condition is true and 0 otherwise.

During the period of control being taken over, we additionally maintain the recovery buffer $B_{\text{recovery}}$ of size $R$ to assess whether the EGO vehicle is safe to return control to the ADS. Specifically, for each frame $t$, a value of 1 is recorded in the buffer if any potential hazard is detected, otherwise a value of 0 is written, as formulated in Eq.~\ref{recovery}.
\begin{equation}\label{recovery}
    \small
\begin{array}{rr}
B_{\text{recovery}}[t \bmod R] = \mathbb{I}(
 B_{\text{collision}}[t \bmod M] \lor \\
 B_{\text{signal}}[t \bmod M] \lor 
 B_{\text{stall}}[t \bmod N])
\end{array}
\end{equation}


\subsection{Hazard Mitigator}\label{Hazard Mitigator}

To mitigate potential hazards, we adopt a conservative driving strategy based on the IDM, which temporarily takes over the control from the ADS until safety is reestablished. This process involves two key steps, \ie the \textbf{waypoint rerouting}, which generates alternative waypoints to guide the EGO vehicle around obstacles whenever possible, and the \textbf{leading actor augmentation}, which dynamically augments the set~of leading actors used by IDM to account for different types of hazards, thereby regulating the vehicle speed to~ensure~safe~navigation.

\begin{algorithm}[!t] 
    \caption{Waypoint Rerouting}
    \footnotesize
    \label{alg:waypointRerouting}
    \LinesNumbered 
    \KwIn{EGO vehicle's position: $p_E^t$, EGO vehicle's length: $L$, navigation point: $NP$, perception range: $perRange$,\ \ \ \ \ \     
    static obstacles: $SO$, road boundaries: $RB$}
    \KwOut{the rerouted waypoints: $RW$}
    
    $ref\_path \leftarrow$ GenerateBézierCurve$(p_E^t, NP)$\; \label{bezier_begin}
    $dense\_wps \leftarrow$ Sampling$(ref\_path)$\; \label{bezier_end}

    $map \leftarrow$ InitOccupancyMap$(p_E^t, perRange)$\; \label{Grid_begin}
    \ForEach{obstacle or boundary $r \in SO \cup RB$}{\label{Grid_color}
        $expanded\_r \leftarrow$ ExpandOccupancy$(r, L)$  \; \label{Expand}
        \ForEach{cell $c \in expanded\_r$}{
            Mark $c$ as $non\_traversable$ \;
        }
    }\label{Grid_end}

    $RW \leftarrow [p_E^t]$\;\label{RW_begin}
    \ForEach{waypoint $wp$ in $dense\_wps$}{\label{RW_for}
        $wp\_cell \leftarrow$ GetCellFromMap($wp$, $map$)\; \label{cell_lookup}
        \If{$wp\_cell$ is traversable}{
            Add $wp$ to $RW$\;
        }\label{RW_if}
        \Else{
            $start \leftarrow$ last waypoint in $RW$\;\label{RW_start}
            $goal \leftarrow$ FindNextTraversableCell($wp$, $dense\_wps$, $map$)\;\label{RW_goal}
            $local\_path \leftarrow$ A*($start$, $goal$, $map$) with penalties\;\label{RW_Astar}
            Append $local\_path$ to $RW$\;\label{RW_append}
        }
    }
    $RW \leftarrow$ Smooth($RW$)\; \label{RW_end}
    \Return{$RW$}\; \label{end}

\end{algorithm}

\begin{figure}[!t]
    \centering
    \subfloat[Dense Waypoints]{%
        \includegraphics[width=0.31\linewidth]{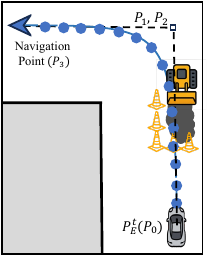}%
        \label{img:WR:a}}
    \hspace{0.0025\linewidth}
    \subfloat[Occupancy Map]{%
        \includegraphics[width=0.31\linewidth]{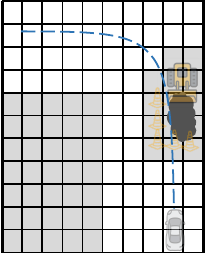}%
        \label{img:WR:b}}
    \hspace{0.0025\linewidth}
    \subfloat[Rerouted Waypoints]{%
        \includegraphics[width=0.31\linewidth]{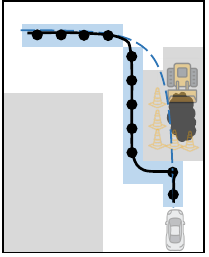}%
        \label{img:WR:c}}

    \caption{An Example of Waypoint Rerouting}
    \label{img:WR}

\end{figure}

\textbf{Waypoint Rerouting.} At each frame $t$ during the takeover process, we generate a set of guiding waypoints $RW$ for the EGO vehicle based on its current position $p_E^t$, its length $L$ and the navigation point $NP$ provided by the ADS, and the perception range $perRange$, static obstacles $SO$ along with the road boundary $RB$ extracted from the BEV representation $BEV^t$. The overall process is illustrated in Algorithm~\ref{alg:waypointRerouting},~which consists of three main steps, \ie \textit{Dense Waypoint Generation}~(Line~\ref{bezier_begin}-\ref{bezier_end}), \textit{Occupancy Map Generation} (Line~\ref{Grid_begin}-\ref{Grid_end}), and \textit{Rerouted Waypoint Generation} (Line~\ref{RW_begin}-\ref{RW_end}). Additionally, Fig.~\ref{img:WR} shows an example of the waypoint rerouting, where the gray regions denote non-traversable areas (\ie road boundaries or static obstacles), and the blue regions show the rerouted waypoints.

\textit{Dense Waypoint Generation.} First, we construct a reference path $ref\_path$ that connects $p_E^t$ and $NP$ (Line~\ref{bezier_begin}). 
Similar to previous work~\cite{feng2022rethinking, liu2021end}, 3-rd Bézier curve \cite{mortenson1999mathematics}, which is empirically found  sufficient for modeling lane lines, is used to compute a smooth reference path for this phase. 
A 3-rd Bézier curve $\mathcal{B}$ can be constructed by four control points $P_{0}-P_{3}$, i.e., $\mathcal{B}(\zeta)=(1-\zeta)^{3}P_{0}+3(1-\zeta)^{2}\zeta P_{1}+3(1-\zeta)\zeta^{2}P_{2}+\zeta^{3}P_{3},\ \zeta \in [0,1]$.
Next, we sample points along this reference path to obtain a set of~dense waypoints (Line~\ref{bezier_end}). 
For example, as shown in Fig.~\ref{img:WR:a}, the EGO vehicle is stalled due to a construction zone and the ADS fails to generate a valid trajectory to bypass the obstacle. We set the $p_E^t$ as $P_0$, and the $NP$ as $P_3$.
Then, we compute the intermediate control point $P_1$ and $P_2$, located~at the intersection of the heading lines of the EGO vehicle and the navigation point within the drivable area to guide the curve plausibly.
Using these four points, we construct a curve and sample it to obtain a set of dense waypoints.
Note that the dense waypoints may not always result in a drivable path, especially in the presence of dynamic or complex obstacles. 
Therefore, the rerouted waypoint generation and the leading actor augmentation work together to ensure safe driving control.

\textit{Occupancy Map Generation.} After the generation of dense waypoints, we initialize a local occupancy grid map centered on the EGO vehicle. Specifically, an occupancy map is initialized using the current position of the EGO vehicle as the origin and a predefined perception range, covering the surrounding area likely to influence path planning decisions (Line~\ref{Grid_begin}). 
To enable fine-grained path search, we adopt a small grid cell size (e.g., 1.0\,m) when generating the occupancy map, which is consistent with practices in robotic navigation~\cite{koenig2005fast,hart1968formal}.~However, unlike robots whose physical footprint is often smaller than a grid cell, vehicles are significantly larger. This discrepancy necessitates additional considerations to ensure safe path planning.
We inflate each static obstacle by half the length of the EGO vehicle (Line~\ref{Expand}), ensuring no collision occurs between the EGO vehicle and nearby obstacles during waypoint rerouting. Each cell that overlaps with the obstacles or road boundaries is marked as non-traversable (Line~\ref{Grid_color}-\ref{Grid_end}), forming a binary occupancy map that encodes drivable and non-drivable regions for subsequent use in rerouting.
For example, as~shown in Fig.~\ref{img:WR:b}, the occupancy map is initialized with a perception range of 40\,m. For clarity, only a local part of the occupancy map is visualized, where the cells overlapping with the~construction zone and road boundary are marked as non-traversable.

\textit{Rerouted Waypoint Generation.} Using the previously constructed occupancy map, we refine the dense waypoints to safely circumvent obstacles. Starting from the $p_E^t$ (Line~\ref{RW_begin}), we sequentially append each dense waypoint to the rerouted waypoint list $RW$, unless a waypoint is found to be located in a non-traversable cell of the occupancy map (Line~\ref{RW_for}-\ref{RW_if}).
The cell corresponding to the last waypoint in the current $RW$ is recorded as a new $start$ point (Line~\ref{RW_start}), indicating where the original waypoints start to become invalid due to environmental constraints. 
We use the function \textit{FindNextTraversableCell} to iterate through and remove waypoints that are in non-traversable cells in dense waypoints until it finds a waypoint located in a traversable cell, which is then used as the new $goal$ point (Line~\ref{RW_goal}).
Then, from the $start$ point, the remaining portion of the dense waypoints is replanned using an A*~\cite{hart1968formal} search algorithm, initialized with the $start$ as the start node and the $goal$ as the target (Line~\ref{RW_Astar}). 
To encourage safer and more efficient detours, we additionally integrate cost penalties for deviations from the reference dense waypoints and for abrupt directional changes into the A* algorithm. These penalties collectively guide the search process to favor not only the shortest path but also one that conforms to safety and motion feasibility considerations.
The resulting path segment is appended to $RW$, producing a complete, collision-free waypoint sequence (Line~\ref{RW_append}). 
After appending the rerouted path segment, the algorithm resumes iterating through the remaining dense waypoints, repeating the process as above.
Finally, a smoothing operation is applied to eliminate abrupt turns or jagged transitions before returning the rerouted waypoints, ensuring the continuity~and drivability (Line~\ref{RW_end}-\ref{end}).
Fig.~\ref{img:WR:c} illustrates the rerouted waypoints we generate to navigate around the construction zone.

\textbf{Leading Actor Augmentation.}
We utilize the IDM to compute the desired speeds of the EGO vehicle along the rerouted waypoints. Following previous work~\cite{treiber2000congested}, we set the desired speed $v_0$ to 0.72 times lane-specific speed limit, $s_0$ to $4.0$m and $T$ to $0.25$s; and the maximum acceleration~$a$ and the braking deceleration $b$ are fixed to $11$m/s$^2$ and $20$m/s$^2$, respectively. In order to regulate the vehicle speed under potential driving hazards, we augment leading actors for the EGO vehicle, from which we derive the parameters (\ie the net distance between the EGO vehicle and the leading actor $ s_E$ and their relative speed $\Delta v_E$) of the IDM introduced in~Sec.~\ref{sec:IDM}.

By default, we select the closest vehicle along the rerouted waypoints in front of the EGO vehicle as a leading actor.
To further enhance safety, if the \textit{Hazard Monitor} detects a potential collision over the prediction horizon $[t, t+H]$, we additionally include those traffic participants as leading actors, such as vehicles and pedestrians, that are predicted to collide with the EGO vehicle within this horizon.
Moreover, static obstacles located along the EGO vehicle's trajectory are also incorporated. These not only help prevent imminent collisions, but also guide the EGO vehicle to decelerate proactively, preserving space for the waypoint rerouting to avoid potential stalling hazards.
When stop signals that may affect the EGO vehicle are present, we augment the set of leading actors with virtual static bounding boxes of the stop signals, as designed in Sec.~\ref{sec:HM}, to restrict the EGO vehicle's speed and ensure it can come to a complete stop within the influence region of stop signals.

Then, we calculate the $s_E$ along with the $\Delta v_E$ of each leading actor. We apply Eq.~\ref{IDM} to each of these leading actors to generate their respective leading speed, and select the minimum speed as the desired speed to regulate the speed of the EGO vehicle. 
Finally, the \textit{Hazard Mitigator} sends the newly generated trajectory, consisting of the rerouted waypoints and their desired speed, to the vehicle controller for hazard mitigation, and continues to regulate the vehicle's trajectory until the control is returned to the ADS by the \textit{Takeover Gate}.


\subsection{\mr{Formal Guarantees}}\label{Formal Guarantees}

{
As defined in Eq.~\ref{formal}, \tool guarantees that the system is either in a safe state $\Phi_{\text{safe}}$ under the control of the \textit{Hazard Mitigator} (HM) at frame $t$, or remains safe at frame $t$ under the control of the ADS or is able to recover to a safe state at the next frame $t+1$ when the takeover $\mathcal{T}$ is necessary.
\begin{equation}\label{formal}
\small
\begin{array}{ll}
\left [\mathcal{C}_t = \text{HM} \land s_t \in \Phi_{\text{safe}} \right ] \lor \\
\left [\mathcal{C}_t = \text{ADS} \land ((\neg \mathcal{T} \land s_{t} \in \Phi_{\text{safe}}) \lor  (\mathcal{T} \land s_{t+1} \in \Phi_{\text{safe}})) \right ]
\end{array}
\end{equation}
where $\mathcal{C}_t$ denotes the ownership of system control at frame $t$, and $s_t$ denotes the system state at frame $t$. 

This safety guarantee is supported by four key proofs. (a) The initial system state $s_0 \in \Phi_{\text{safe}}$. (b) Given accurate perception and dynamics, our \textit{Hazard Monitor} detects all hazards within our defined hazard set. (c) If $\mathcal{C}_t = \text{HM}$, our IDM-based \textit{Hazard Mitigator} mathematically guarantees no rear-end collision with any leading actors and no stalling, ensuring $\Phi_{\text{safe}}$ is forward-invariant, \ie $\forall k \geq 0, s_{t+k} \in \Phi_{\text{safe}}$. (d) If $\mathcal{C}_t = \text{ADS}$ and the takeover is necessary, our deterministic \textit{Takeover Gate} switches~control within a latency significantly shorter than a single frame, ensuring $s_{t+1} \in \Phi_{\text{safe}}$ based on~(c).
}

\section{Evaluation}
We have implemented a prototype of \tool with 6,709 lines of Python code. The source code and data of our work are available at~\cite{website1}.
To evaluate the effectiveness and efficiency of \tool, we design the following \mr{six} research questions.

\begin{itemize}[leftmargin=*]
    \item \textbf{RQ1 Effectiveness Evaluation}: How effective is \tool in enhancing the resilience of end-to-end ADSs?
    \item \textbf{RQ2 Efficiency Evaluation}: How efficient is \tool in enhancing the resilience of end-to-end ADSs?
    \item \textbf{RQ3 Accuracy Evaluation}: How accurate is \tool in producing appropriate takeover decisions?
    \item \textbf{RQ4 Ablation Study}: How do the steps of \textit{Waypoint Rerouting} and \textit{Leading Actor Augmentation} impact the IDM-based hazard mitigation?
    \item \mr{\textbf{RQ5 Parameter Sensitivity Analysis}: How robust is \tool in different parameter settings?}
    \item \mr{\textbf{RQ6 Generalization Evaluation}: How effective is \tool when generalized to modular ADSs?}
\end{itemize}

\subsection{Evaluation Setup}

\textbf{Benchmark.} To train and evaluate the resilience of the ADS under different driving hazards, we select Bench2Drive~\cite{jia2024bench} and CARLA leaderboard 2.0 validation set~\cite{CARLA2.0}, which are widely used benchmarks containing various driving hazards. Bench2Drive's training set consists of 2 million fully annotated frames, collected from 12 towns under 23 weathers by the expert model~\cite{li2024think} in CARLA~\cite{Dosovitskiy17}. Its evaluation set requires the ADS to pass 44 interactive scenarios under different locations and weathers,  which sums up to 220 routes (denoted as Bench2Drive220). Besides, the CARLA leaderboard 2.0 validation set contains 20 long routes with different weather and traffic conditions (denoted as CLV20). The closed-loop driving task requires the ADS to drive toward the destination point under different driving hazards. 

\textbf{Target ADS.} We choose three state-of-the-art models (\ie TCP~\cite{wu2022trajectoryguided}, UniAD~\cite{hu2023planning} and VAD~\cite{jiang2023vad}), which are the most~representative end-to-end ADSs, as our targets. All~the three ADSs are initially trained with the Bench2Drive training~set. 

\textbf{Resilience Metrics.} First, to measure the overall effectiveness of the ADS under driving hazards, we select the metrics named \textit{Success Rate } ($SR$), \textit{Route Completion} ($RC$) and \textit{Driving Score} ($DS$)~\cite{Dosovitskiy17}. $SR$ measures the proportion of routes that are successfully completed within the allotted time and without any violations. $RC_i$ denotes the completion ratio of route $i$, while $RC$ represents the average completion ratio across all routes. $DS$ is defined by Eq.~\ref{eq:driving_score},
\begin{equation}\label{eq:driving_score}
    \begin{array}{c}
        DS = \frac{1}{R_{total}} \sum_{i=1}^{R_{total}} RC_{i}\times IS_{i}
    \end{array}
\end{equation}
where $IS_{i}$ is a penalty factor generated by the evaluation benchmark, starting at 1.0, which gets reduced with each~violation (\eg vehicle collision), and $R_{total}$ is the total number of routes. 

Second, to evaluate the multiple skills of ADSs under driving hazards, the evaluation routes in Bench2Drive are officially categorized into five groups~\cite{jia2024bench},~\ie merging ($Merge$), overtaking ($Overtake$), emergency braking ($EmgBrake$), giving way ($GiveWay$) and obeying traffic signs ($TSign$), for which success rates are reported per~category. 

Finally, to evaluate the capability of the ADS to prevent violations in hazardous scenarios, we quantify the number of violations per kilometer from the testing reports provided by CARLA across three categories, \ie vehicle collision ($Coll$), failure to obey stop signals ($Stop$), and vehicle stalling ($Stall$).

\textbf{Research Question Setup.} For \textbf{RQ1}, \textbf{RQ2} and \textbf{RQ3}, we first bridge the three base models (\ie TCP, UniAD and VAD) with CARLA, using all routes (\ie 240 routes in total) in Bench2Drive220 and CLV20. Then, we adopt a widely used approach, \ie BEVFormer~\cite{BEVFormer}, which aggregates multi-view camera inputs to construct the surrounding environment of the ADS. We integrate \tool with the three base models based on BEVFormer perception, resulting in \TCPtools, \UniADtools, and \VADtools, respectively.
In addition, following previous work~\cite{DianChen, wang2025adawm, gao2024enhance, li2024think}, we use the privileged environment information provided by CARLA, such as surrounding vehicle states and stop signals' location, to construct the BEV representations. We integrate \tool with these three base models~using the privileged BEVs, and refer to these instantiated variants as \TCPtool, \UniADtool and \VADtool, respectively. 
All the six variants are evaluated on the same set of 240 routes, to assess the effectiveness and efficiency of \tool under conditions of realistic perception-based BEVs and the ideal privileged environment information,~respectively. 

For \textbf{RQ1}, we first report the average resilience metrics of the base models and their \tool-enhanced variants on two benchmarks. Then, we conduct an in-depth analysis of \tool's effectiveness in representative hazardous scenarios.

For \textbf{RQ2}, to measure the efficiency of \tool, we report the average wall-clock time of each frame in the simulator taken by the three components in \tool (\ie the \textit{Takeover Gate}, the \textit{Hazard Monitor} and the \textit{Hazard Mitigator}), separately.

For \textbf{RQ3}, following~\cite{stocco2022thirdeye}, we leverage the violation timestamps provided by CARLA, and define a takeover occurring within 3 seconds prior to a violation as a necessary intervention. Based on this, we count the number of necessary takeovers as true positives (TP), unnecessary takeovers as false positives (FP), and violations that were not preceded by a takeover as false negatives (FN).
Our primary goal is to achieve high recall, defined as Recall = TP / (TP + FN), because the cost associated with false negatives is very high in the safety-critical domain~\cite{blair1979information}, as they may lead to violations.
Achieving high precision, defined as Precision = TP / (TP + FP), is also important to avoid unnecessary interventions by \tool.
Following previous work~\cite{stocco2022thirdeye}, we report the $\text{F}_{\beta}$ score~\cite{blair1979information}, which is a weighted harmonic mean of precision and recall, with $\beta = 3$ to emphasize recall over precision.

For \textbf{RQ4}, we conduct ablation studies to evaluate the individual contribution of the \textit{Waypoint Rerouting} and the \textit{Leading Actor Augmentation} steps within \tool. To this end, we implement three reduced variants of \tool, \ie one without \textit{Waypoint Rerouting}, one without \textit{Leading Actor Augmentation}, and one without both steps (\ie the IDM-only approach). Since evaluating a single ADS using all 240 routes takes at least 5 days, we follow~\cite{jia2025drivetransformer} to conduct studies using Dev10~\cite{Dev10}, a subset of Bench2Drive220 containing 10 routes, specifically designed to reduce evaluation variance and save computational resources. We evaluate these variants using realistic perception-based BEVs on Dev10 and compare them with their full counterparts (\ie \TCPtools, \UniADtools, and \VADtools) to focus on the average resilience~metrics.

\mr{For \textbf{RQ5}, we conduct parameter sensitivity analysis using realistic perception-based BEVs by varying buffer settings (\ie collision/stalling buffer length $M$, threshold $l$, and recovery buffer length $R$), and bounding box enlargement factors. We evaluate these variants on Dev10 to focus on the average accuracy and resilience~metrics.}

\mr{For \textbf{RQ6}, we integrate \tool into Apollo 7.0, a widely used open-source modular ADS, and bridge it with CARLA~following~\cite{carla_apollo_bridge}. We evaluate the effectiveness on Bench2Drive220 and CLV20, comparing \tool with REDriver~\cite{sun2024redriver}, a~state-of-the-art safety assurance approach for Apollo by refining trajectories with signal temporal logic to avoid violating traffic~rules.}

To mitigate the impact of randomness, all the experiments are conducted five times, and the average results are reported.

\textbf{Environment.} We conduct all the experiments on Ubuntu 20.04.4 LTS servers with 4 NVIDIA GeForce RTX 3090 GPUs, Intel(R) Xeon(R) Silver 4310 @ 2.10GHz and 128GB memory.


\subsection{Effectiveness Evaluation (RQ1)}

\begin{table}[!t]
    \caption{Results of the Overall Effectiveness}
    \vspace{-5pt}
    \label{rq1:table1}
    \centering
    \begin{adjustbox}{width=0.9\linewidth}
    \begin{tabular}{lcccccc}
        \toprule
        \multirow{2}{*}{\textbf{Approach}} & \multicolumn{3}{c}{\textbf{Bench2Drive220}} & \multicolumn{3}{c}{\textbf{CLV20}} \\
        \cmidrule(lr){2-4} \cmidrule(lr){5-7}
         & $SR\uparrow$ & $RC\uparrow$ & $DS\uparrow$ & $SR\uparrow$ & $RC\uparrow$ &$DS\uparrow$ \\
        \midrule
        \textbf{TCP} & 14.55 & 51.53 & 39.28 & 0.00 & 2.21 & 0.84\\
        \rowcolor[gray]{0.9} \textbf{\TCPtools} & 32.73 & 72.09 & 60.18 & 0.00 & 4.78 & 1.56\\
        \rowcolor[gray]{0.9} \textbf{\TCPtool} & 38.64 & 76.02 & 64.72 & 0.00 & 4.81 & 2.03\\
        \midrule
        \textbf{UniAD} & 14.09 & 68.68 & 44.62 & 0.00 & 2.05 & 0.50 \\
        \rowcolor[gray]{0.9} \textbf{\UniADtools} & 39.55 & 83.95 & 68.61 & 0.00 & 9.38 & 2.11 \\
        \rowcolor[gray]{0.9} \textbf{\UniADtool} & 52.27 & 89.56 & 75.90 & 0.00 & 11.56 & 2.27\\
        \midrule
        \textbf{VAD} & 17.27 & 61.60 & 43.31 & 0.00 & 0.93 & 0.37\\
        \rowcolor[gray]{0.9} \textbf{\VADtools} & 39.09 & 83.66 & 66.98 & 0.00 & 6.15 & 1.60 \\
        \rowcolor[gray]{0.9} \textbf{\VADtool} & 44.55 & 85.95 & 71.51 & 0.00 & 6.79 & 2.09  \\
        \bottomrule
    \end{tabular}
\end{adjustbox}
\end{table}

\textbf{Overall Effectiveness.} 
As shown in Table~\ref{rq1:table1}, with respect to Bench2Drive220, \tool improves $SR$ by 144.00\%, $RC$ by 32.65\%, and $DS$ by 53.88\%, on average, across three ADSs. 
When the privileged environment information is available, \tools achieves even greater improvements of 198.17\% in $SR$, 39.15\% in $RC$, and 66.66\% in $DS$. With respect to CLV20, we observe that $SR$ remains 0 across all target ADSs and their \tool-enhanced variants. This is because the routes in CLV20 are much longer and the hazard scenarios are more complex, making it difficult for the ADSs to fully complete the driving task.  Despite this, \tool and \tools exhibit improvements in both $RC$ and $DS$ across all target ADSs. Specifically, \tool achieves average improvements of 345.05\% in $RC$ and 246.72\% in $DS$, while \tools leads to improvements of 403.89\% in $RC$ and 320.18\% in $DS$. 

To sum up, \tool and \tools demonstrate superior improvements across all target ADSs on Bench2Drive220 and CLV20. Compared to \tools, which leverages privileged environment information, \tool achieves comparable gains with less reliance on ground-truth data, and its average improvements on $SR$, $RC$, and $DS$ are only 4.02, 2.45, 2.91 lower, respectively. These results underscore the value of \tool in enhancing resilience, significantly extending the overall effectiveness of target ADSs in the face of hazards.

\begin{table}[!t]
    \caption{Results of the Multi-Skill Performance}
    \vspace{-5pt}
    \label{rq1:table2}
    \centering
    \begin{adjustbox}{width=0.9\linewidth}
    \begin{tabular}{lccccc}
        \toprule
        \multirow{2}{*}{\textbf{Approach}} & \multicolumn{5}{c}{\textbf{Bench2Drive220}} \\
        \cmidrule(lr){2-6}
         & $Merge\uparrow$ & $Overtake\uparrow$ & $EmgBrake\uparrow$ & $GiveWay\uparrow$ & $TSign\uparrow$  \\
        \midrule
        \textbf{TCP} & 14.29 & 20.00 & 20.00 & 10.00 & 5.91 \\
        \rowcolor[gray]{0.9} \textbf{\TCPtools} & 34.25 & 20.00 & 45.00 & 30.00 & 41.40 \\
        \rowcolor[gray]{0.9} \textbf{\TCPtool} & 36.36 & 33.33 & 46.67 & 40.00 & 44.21 \\
        \midrule
        \textbf{UniAD} & 12.66 & 13.33 & 20.00 & 10.00 & 13.23 \\
        \rowcolor[gray]{0.9} \textbf{\UniADtools} & 29.11 & 22.22 & 65.00 & 50.00 & 50.53 \\
        \rowcolor[gray]{0.9} \textbf{\UniADtool} & 36.25 & 51.11 & 76.67 & 50.00 & 57.89 \\
        \midrule
        \textbf{VAD} & 13.51 & 20.00 & 25.42 & 20.00 & 23.66 \\
        \rowcolor[gray]{0.9} \textbf{\VADtools} & 31.25 & 20.00 & 63.33 & 30.00 & 51.58 \\
        \rowcolor[gray]{0.9} \textbf{\VADtool} & 31.65 & 44.44 & 63.33& 50.00 & 53.19  \\
        \bottomrule
    \end{tabular}
\end{adjustbox}
\end{table}

\textbf{Multi-Skill Performance.} 
As shown in Table~\ref{rq1:table2}, \tool and \tools significantly enhance the multiple skills of the target ADSs under driving hazards on Bench2Drive220.
\tool achieves an average improvement of 174.48\% on the five driving skills across all target ADSs, with the most significant improvement observed in $TSign$ (333.48\%) and the least in $Overtaking$ (22.23\%). Besides, \tools achieves an average improvement of 231.57\% across all skills, with the most significant improvement observed in $TSign$ (370.14\%) and the least in $Overtaking$ (157.42\%). Overall, even when relying solely on perceived environmental information, \tool effectively enhances the driving skills of the three ADSs, demonstrating its broad applicability and strong generalization~capability.

\begin{table}[!t]
    \caption{Results of the Violation Prevention}
    \vspace{-5pt}
    \label{rq1:table3}
    \centering
    \begin{adjustbox}{width=0.9\linewidth}
    \begin{tabular}{lcccccc}
        \toprule
        \multirow{2}{*}{\textbf{Approach}} & \multicolumn{3}{c}{\textbf{Bench2Drive220}} & \multicolumn{3}{c}{\textbf{CLV20}} \\
        \cmidrule(lr){2-4} \cmidrule(lr){5-7}
         & $Coll\downarrow$ & $Stop\downarrow$ & $Stall\downarrow$ & $Coll\downarrow$ & $Stop\downarrow$ & $Stall\downarrow$ \\
        \midrule
        \textbf{TCP} & 8.23 & 0.21 & 3.51 & 4.64 & 0.17 & 2.5 \\
        \rowcolor[gray]{0.9} \textbf{\TCPtools} & 4.31 & 0.17 & 0.64 & 3.01 & 0.08 & 0.75\\
        \rowcolor[gray]{0.9} \textbf{\TCPtool} & 4.19 & 0.00 & 0.55 & 1.65 & 0.00 & 0.74 \\
        \midrule
        \textbf{UniAD} & 10.66 & 1.51 & 2.30 & 11.17 & 2.46 & 3.32\\
        \rowcolor[gray]{0.9} \textbf{\UniADtools} & 4.90 & 0.20 & 0.61 & 2.33 & 0.22 & 0.79 \\
        \rowcolor[gray]{0.9} \textbf{\UniADtool} & 3.90 & 0.00 & 0.48 & 2.21 & 0.00 & 0.63 \\
        \midrule
        \textbf{VAD} & 10.67 & 0.41 & 3.11 & 16.09 & 0.42 & 8.47 \\
        \rowcolor[gray]{0.9} \textbf{\VADtools} & 5.16 & 0.26 & 0.92 & 2.59 & 0.20 & 1.22 \\
        \rowcolor[gray]{0.9} \textbf{\VADtool} & 4.14 & 0.00 & 0.44 & 2.22 & 0.00 & 1.05\\
        \bottomrule
    \end{tabular}
\end{adjustbox}
\end{table}

\textbf{Violation Prevention.} As shown in Table~\ref{rq1:table3}, both \tool and \tools substantially prevent violations across all target ADSs on Bench2Drive220 and CLV20.
On average across all target ADSs on Bench2Drive220, \tool reduces $Coll$, $Stop$, and $Stall$ by 51.10\%, 51.44\% and 75.22\%, respectively, while \tools further improves these reductions to 57.06\%, 100.00\% and 83.10\%, respectively.  On CLV20, \tool reduces $Coll$, $Stop$, and $Stall$ by 66.06\%, 65.46\%, and 77.02\%, while \tools achieves 76.95\%, 100.00\%, and 79.48\%, respectively.
\tool achieves comparable gains to \tools on the two benchmarks in preventing violations in hazardous scenarios. Notably, \tools achieves perfect compliance with all stop signals under the privileged environment information, fully eliminating the failure to obey stop signals.




\begin{figure*}[!t]
    \centering
    \begin{minipage}[t]{0.9\textwidth}
        \begin{minipage}[t]{0.86\textwidth}
            \centering
            \subfloat[Vehicle Collision]{%
                \includegraphics[width=0.29\textwidth]{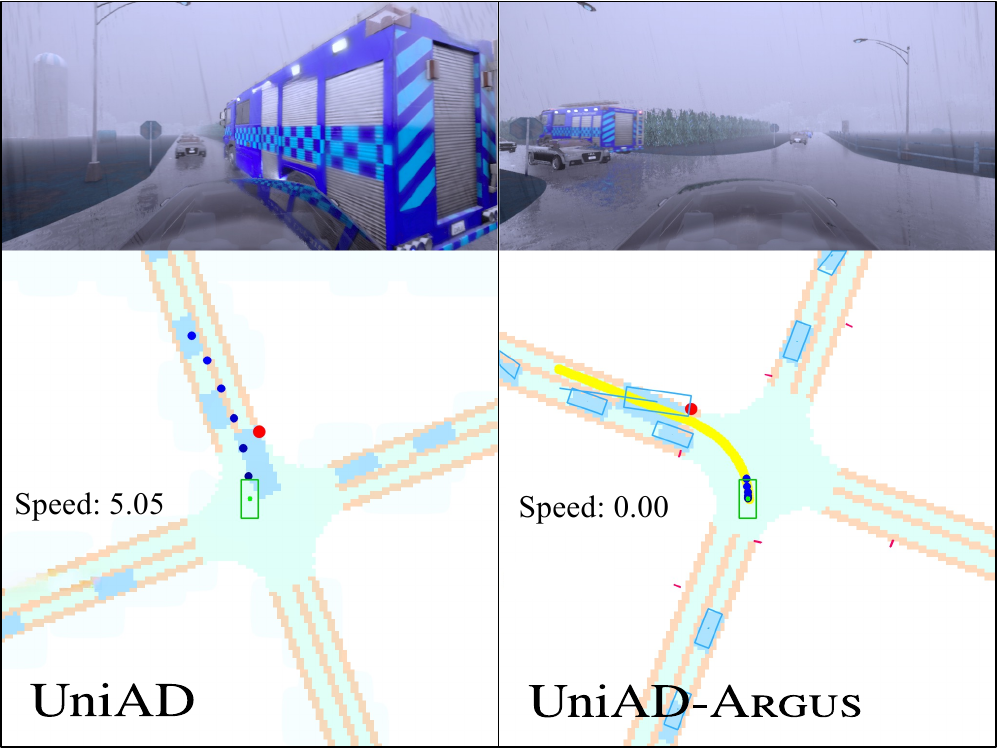}%
                \label{fig:rq1:a}}
            \subfloat[Pedestrian Collision]{%
                \includegraphics[width=0.29\textwidth]{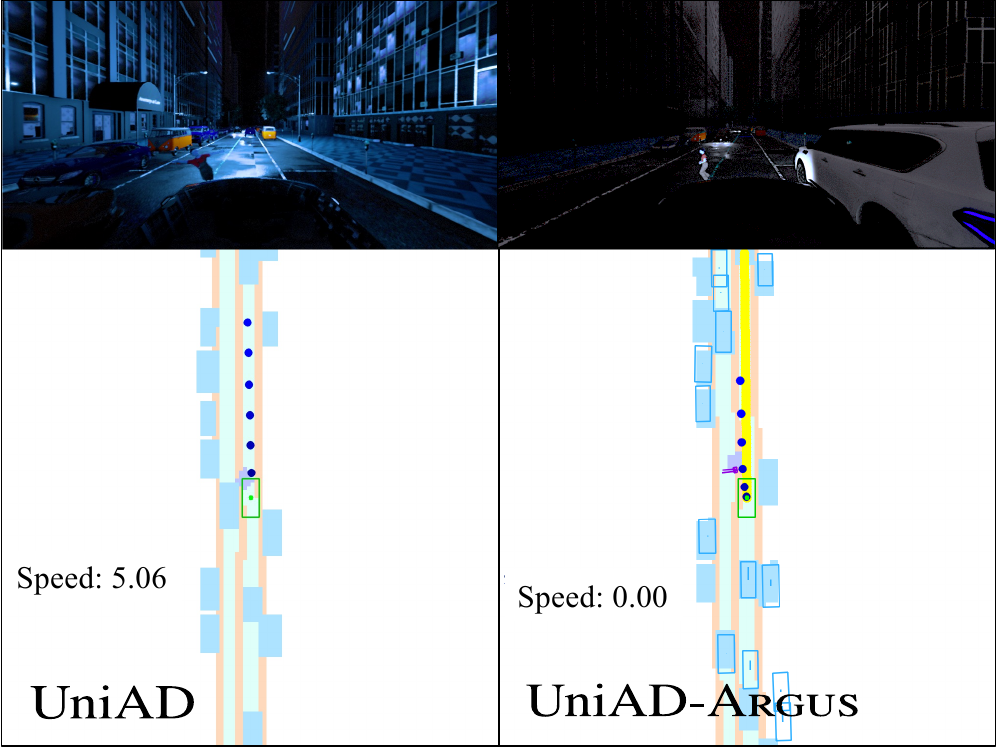}%
                \label{fig:rq1:b}}
            \subfloat[Stop Signal Violation]{%
                \includegraphics[width=0.29\textwidth]{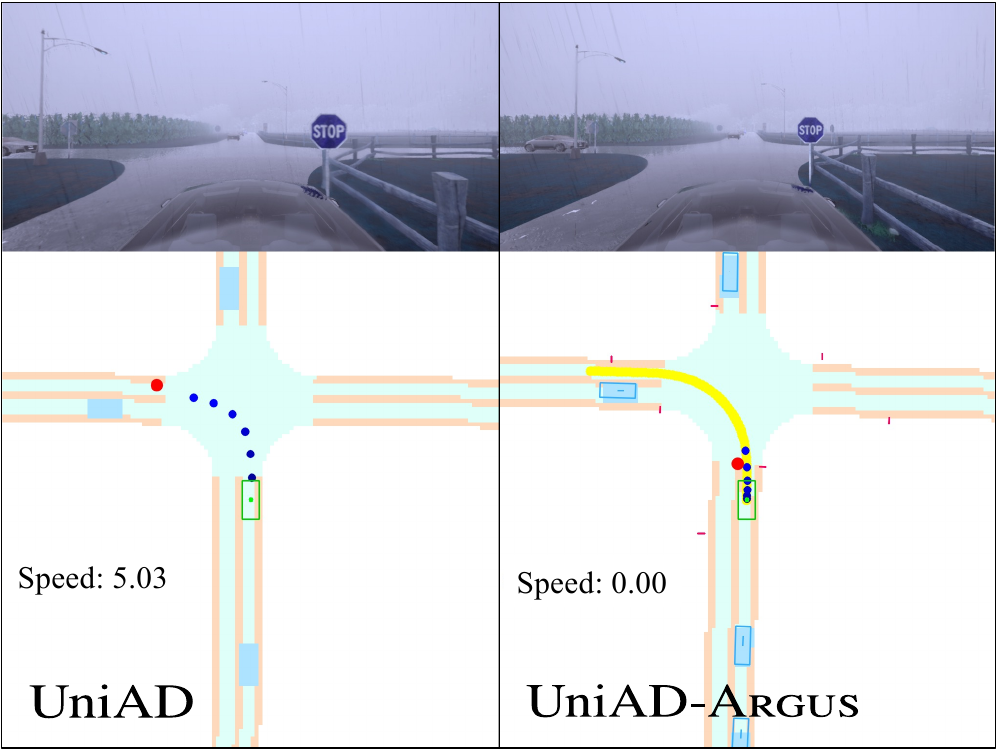}%
                \label{fig:rq1:c}}
            \subfloat[Vehicle Stalling]{%
                \includegraphics[width=0.29\textwidth]{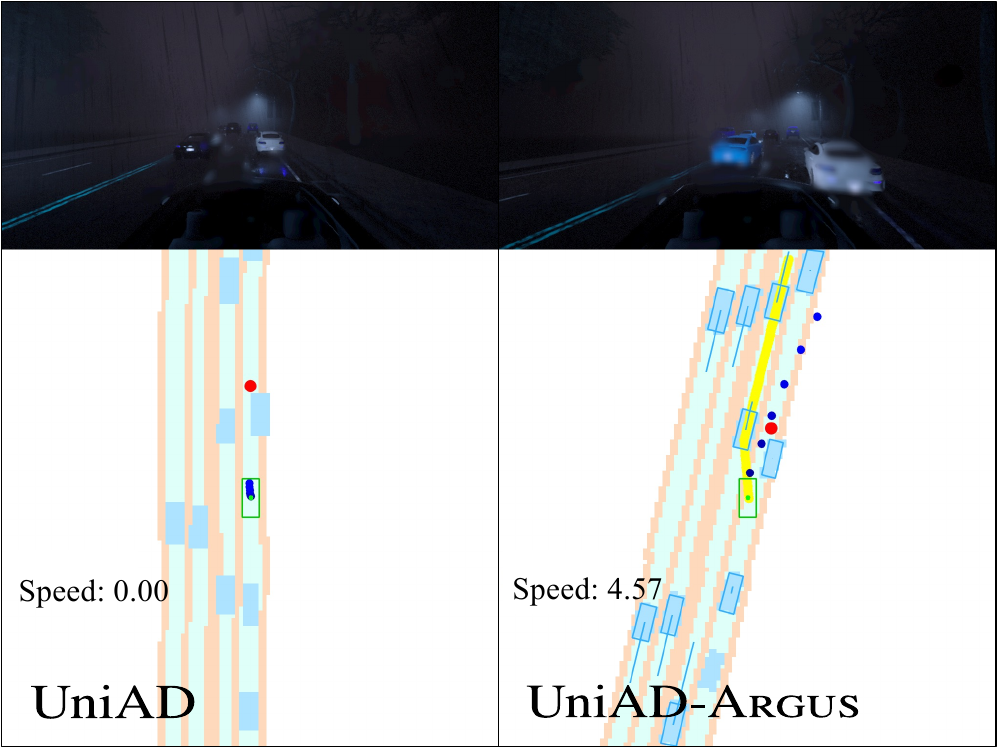}%
                \label{fig:rq1:d}}
        \end{minipage}%
    \end{minipage}
    \includegraphics[width=.8\textwidth]{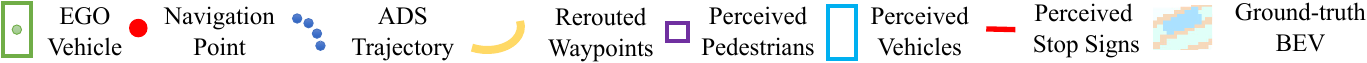}
    \vspace{-3pt}
    \caption{Qualitative Examples of the \UniADtools}
    \label{fig:qualitative examples}
\end{figure*}

\textbf{Case Study.} Fig.~\ref{fig:qualitative examples} illustrates the effectiveness of \tool on UniAD in four representative driving hazards, whose videos are available at our replication website~\cite{website1}.
\begin{itemize}[leftmargin=*]
    \item In Fig.~\ref{fig:rq1:a}, the EGO vehicle encounters a large emergency vehicle running a red light at a high speed under rainy and low-visibility conditions. UniAD fails to react appropriately and proceeds at a speed of 5.05 m/s, resulting in a collision. In contrast, \UniADtools correctly monitors the hazard and takes over in time, bringing the vehicle to a full stop (a speed of 0.00 m/s) and thereby avoiding the collision.
    \item In Fig.~\ref{fig:rq1:b}, the EGO vehicle encounters a pedestrian emerging from behind a parked vehicle and entering the lane at night. UniAD fails to generate a safe trajectory and adjust its speed in time, leading to a collision with the pedestrian at a speed of 5.06 m/s. In contrast, \UniADtools successfully monitors the hazard, takes over control, and stops the vehicle before reaching the pedestrian, thereby avoiding the collision.
    \item In Fig.~\ref{fig:rq1:c}, the EGO vehicle approaches an intersection controlled by a clearly visible stop sign. UniAD fails to perform the mandatory stop and continues through the intersection at a speed of 5.03 m/s, resulting in a stop sign violation, while \UniADtools monitors the potential hazard and takes over to enforce adherence to traffic rules.
    \item In Fig.~\ref{fig:rq1:d}, the EGO vehicle encounters a parked vehicle partially obstructing the lane, and must change lanes into same-direction traffic. UniAD fails to plan a feasible trajectory and stalls indefinitely (a speed of 0.00 m/s), while \UniADtools takes over, reroutes around the obstruction, and safely guides the lane change. Control is then smoothly returned to the ADS, allowing driving to resume without~interruption.
\end{itemize}

    \textit{\textbf{Summary.}}  \tool can effectively enhance the resilience of end-to-end ADSs in the face of driving hazards, significantly improving their overall effectiveness, multiple driving skills, and the capability to prevent violations.


\subsection{Efficiency Evaluation (RQ2)}

\begin{table}[!t]
    \caption{Results of Efficiency of Each Module (ms)}
    \vspace{-5pt}
    \label{rq2:table1}
    \centering
    \begin{adjustbox}{width=0.9\linewidth}
    \begin{tabular}{lcccccc}
        \toprule
        \textbf{Approach} & \textit{Takeover Gate} & \textit{Hazard Monitor} & \textit{Hazard Mitigator} \\
        \midrule
        \textbf{\TCPtools} & 3e-03 & 363.45 & 12.24\\
        \textbf{\TCPtool} & 3e-03 & 3.94 & 12.80\\
        \midrule
        \textbf{\UniADtools} & 3e-03 & 364.43 & 12.19\\
        \textbf{\UniADtool} & 3e-03 & 3.78 & 12.16\\
        \midrule
        \textbf{\VADtools} & 3e-03 & 368.31 & 12.13\\
        \textbf{\VADtool} & 3e-03 & 3.82 & 12.24\\
        \bottomrule
    \end{tabular}
\end{adjustbox}
\end{table}

Table~\ref{rq2:table1} reports the average wall-clock time of each frame in the simulator taken by the three components in \tool. With respect to the \textit{Takeover Gate} in \tool and \tools, the average costs across all three target ADSs are both 3e-03 ms, which is negligible compared to the overall runtime of the target ADS. 
This is because the \textit{Takeover Gate} is integrated into the trajectory flow of the target ADS as a gating mechanism, which merely checks the takeover and recovery buffers, thereby introducing no measurable runtime overhead to the target ADS.

With respect to the \textit{Hazard Monitor} in \tool, the average cost across all three target ADSs is 365.40 ms. This latency is primarily attributed to the use of BEVFormer for aggregating multi-view camera inputs to construct the surrounding environment. 
However, the \textit{Hazard Monitor} operates in parallel with the target ADS and computes faster than the target ADS's inference speed (\eg average 534.61 ms per frame for UniAD), ensuring that every predicted trajectory can be monitored without affecting the target ADS's real-time performance.
When using the privileged environment information, the average latency of the \textit{Hazard Monitor} in \tools is reduced to only 3.85 ms, proving that the additional cost of the \textit{Hazard Monitor} is primarily due to the BEV representation construction, which can be further optimized in future work.

With respect to the \textit{Hazard Mitigator} across all target~ADSs, \tool consumes a comparable average runtime of 12.18~ms to \tools, which consumes an average of 12.40 ms. Given that typical end-to-end ADSs require several hundred milliseconds per inference, this additional cost accounts for~only approximately 5\% average overhead across the three target~ADSs, which is practical for mitigating driving hazards in real time.

    \textit{\textbf{Summary.}} \tool can efficiently enhance the resilience of end-to-end ADSs and each component in \tool incurs little additional time overhead.


\subsection{Accuracy Evaluation (RQ3)}

\begin{table}[!t]
    \caption{Accuracy of the \textit{Hazard Monitor}}
    \vspace{-5pt}
    \label{rq3:table1}
    \centering
    \begin{adjustbox}{width=0.9\linewidth}
    \begin{tabular}{lcccccc}
        \toprule
        \multirow{2}{*}{\textbf{Approach}} &  \multicolumn{3}{c}{\textsc{Argus}} & \multicolumn{3}{c}{\textsc{Argus}$^*$} \\
        \cmidrule(lr){2-4} \cmidrule(lr){5-7}
        & Precision & Recall & $\text{F}_3$ & Precision & Recall & $\text{F}_3$ \\
        \midrule
        \textbf{TCP} & 0.639 & 0.964 & 0.917 & 0.652 & 0.976 & 0.930\\
        \textbf{UniAD} & 0.664 & 0.949 & 0.910 & 0.710 & 0.957 & 0.925\\
        \textbf{VAD} & 0.615 & 0.912 & 0.870 & 0.710 & 0.978 & 0.943\\
        \bottomrule
    \end{tabular}
    \end{adjustbox}
\end{table}

Table~\ref{rq3:table1} presents the accuracy of \tool and \tools in producing appropriate takeover decisions. For \tool, the average precision, recall, and $\text{F}_3$ score across all three target ADSs are, on average, 0.639, 0.942, and \todo{0.899}, respectively. When the privileged environment information is utilized by \tools, these metrics improve further to 0.691 in precision, 0.970 in recall, and \todo{0.932} in $\text{F}_3$ score,~on average. 
Due~to~the imprecision of the BEV representation generated by the BEVFormer, the \textit{Hazard Monitor} in \tool tends to miss some necessary takeovers, while producing more false alarms than that in \tools. Nevertheless, its high recall and $\text{F}_3$ score indicate that \tool can still effectively identify the majority of hazardous situations and produce~appropriate~takeover~decisions.

    \textit{\textbf{Summary.}} \tool can produce appropriate takeover decisions in the face of driving hazards.


\subsection{Ablation Study (RQ4)}

\begin{table*}[!t]
    \caption{Ablation Study for the \textit{Hazard Mitigator}}
    \vspace{-5pt}
    \label{rq4:table1}
    \centering
    \begin{adjustbox}{width=0.75\linewidth}
    \begin{tabular}{lccccccccc}
        \toprule
         \multirow{2}{*}{\textbf{Approach}} & \multicolumn{2}{c}{\textbf{Step}} &  \multicolumn{3}{c}{\textbf{Overall}} & \multicolumn{3}{c}{\textbf{Violations}} \\
         \cmidrule(lr){2-3} \cmidrule(lr){4-6} \cmidrule(lr){7-9}
         & \textit{Waypoint Rerouting} & \textit{Leading Actor Augmentation} & $SR\uparrow$ & $RC\uparrow$ & $DS\uparrow$ & $Coll\downarrow$ & $Stop\downarrow$ &$Stall\downarrow$ \\
        \midrule
        \multirow{4}{*}{\textbf{\TCPtools}} & \multicolumn{2}{c}{\mr{IDM-only approach}} & 10.00 & 78.62 & 55.38 & 5.62 & 0.00 & 1.12 \\
        & \ding{51} & & 10.00 & 81.63 & 57.14 & 4.34 & 0.00 & 0.00\\
        & & \ding{51} & 20.00 & 80.93 & 64.61 & 3.30 & 0.00 & 1.10 \\
        \rowcolor[gray]{0.9} & \ding{51} & \ding{51} & 30.00 & 90.61 & 74.87 & 1.00 & 0.00 & 0.00\\
        \midrule
        \multirow{4}{*}{\textbf{\UniADtools}} & \multicolumn{2}{c}{\mr{IDM-only approach}} &10.00 & 88.38 & 52.86 & 12.32 & 1.03 & 1.03 \\
        & \ding{51} & & 20.00 & 91.58 & 59.33 & 8.93 & 0.99  & 0.00\\
        & & \ding{51} &50.00 & 83.70 & 77.42 & 0.00 & 0.00 & 1.07\\
        \rowcolor[gray]{0.9} & \ding{51} & \ding{51} & 60.00 & 98.67 & 83.61 & 0.00 & 0.00 & 0.00\\
        \midrule
        \multirow{4}{*}{\textbf{\VADtools}} & \multicolumn{2}{c}{\mr{IDM-only approach}} & 20.00 & 74.15 & 50.99 & 8.20 & 1.17 & 2.34\\
        & \ding{51} & & 20.00 & 78.67 & 53.43 & 7.77  & 1.11 & 0.00\\
        & & \ding{51} & 50.00 & 72.64 & 69.64 & 0.00 & 0.00 & 1.19 \\
        \rowcolor[gray]{0.9} & \ding{51} & \ding{51} & 60.00 & 92.91 & 83.91 & 0.00 & 0.00 & 0.00\\
        \bottomrule
    \end{tabular}
\end{adjustbox}
\end{table*}

Table~\ref{rq4:table1} presents the results of the ablation study for the \textit{Hazard Mitigator}.
\todo{All reported improvements are based on comparisons with the IDM-only approach.}
Across all target ADSs, the \textit{Waypoint Rerouting} and the \textit{Leading Actor Augmentation} individually contribute significantly to enhancing the resilience of ADSs, improving $DS$ by 6.73\% and 33.23\%, respectively, compared with the IDM-only approach. On one hand, the \textit{Waypoint Rerouting} step generates waypoints to avoid static obstacles, resulting in a 4.51\% increase, on average, in $RC$ and a complete elimination of $Stall$ violations. On the other hand, the \textit{Leading Actor Augmentation} step introduces augmented leading vehicles to guide speed planning, leading to an 80.43\% reduction, on average, in $Coll$ violations and a complete elimination of $Stop$ violations. Besides, when both the two steps are enabled, \tool achieves great improvements in overall effectiveness (\ie a 300\% increase in $SR$, a 17.40\% increase in $RC$, and a 52.64\% increase in $DS$) as well as in violation prevention (\ie a 100\% reduction in $Coll$ and $Stop$ violations, and a 94.07\% reduction in $Stall$ violations). 

    \textit{\textbf{Summary.}} Both the two steps contribute the hazard mitigation, and their combination effectively enhances the resilience of end-to-end ADSs, improving the driving performance.


\subsection{Parameter Sensitivity Analysis (RQ5)}

\begin{table}[!t]
  \caption{\mr{Results of Parameter Sensitivity Analysis}}
  \label{rq4:table0}
  \centering
  \vspace{-5pt}
  \subfloat[Buffer Settings\label{tab:hm_buffer}]{
      \centering
      \begin{adjustbox}{width=.85\linewidth}
      \begin{tabular}{ccccccccc}
        \toprule
        \multirow{2}{*}[-3pt]{$M$} & \multirow{2}{*}[-3pt]{$l$} & \multirow{2}{*}[-3pt]{$R$} & \multicolumn{3}{c}{\textbf{Accuracy}} & \multicolumn{3}{c}{\textbf{Overall}} \\
        \cmidrule(lr){4-6} \cmidrule(lr){7-9}
        & & & Precision & Recall & $\text{F}_3$ & $SR\uparrow$ & $RC\uparrow$ & $DS\uparrow$  \\
        \midrule
        1  & 1 & 20 & 0.300 & 1.000 & 0.811 & 30.00 & 96.41 & 78.49 \\
        3  & 2 & 20 & 0.400 & 1.000 & 0.870 & 40.00 & 92.64 & 72.37 \\
        5  & 3 & 20 & 0.500 & 1.000 & 0.909 & 40.00 & 97.96 & 80.63 \\
        5  & 5 & 20 & 0.571 & 0.667 & 0.656 & 20.00 & 81.35 & 68.72 \\
        10 & 8 & 20 & 0.600 & 0.500 & 0.508 & 20.00 & 80.11 & 64.55 \\
        5  & 4 & 10 & 0.500 & 1.000 & 0.909 & 40.00 & 96.85 & 75.15 \\
        5  & 4 & 40 & 0.667 & 1.000 & 0.952 & 50.00 & 97.29 & 79.73 \\
        \rowcolor[gray]{0.9}5 & 4 & 20 & 0.667 & 1.000 & 0.952 & 60.00 & 98.67 & 83.61\\
        \bottomrule
      \end{tabular}
      \end{adjustbox}
  }\par
  \vspace{-5pt}
  \subfloat[Bounding Box Enlargement Factors\label{tab:hm_boxes}]{
      \centering
      \begin{adjustbox}{width=.98\linewidth}
      \begin{tabular}{ccccccccc}
        \toprule
        \multirow{2}{*}[-3pt]{\makecell{EGO\\Vehicle}} & \multirow{2}{*}[-3pt]{\makecell{Surrounding\\Vehicles}} & \multirow{2}{*}[-3pt]{Pedestrians} & \multicolumn{3}{c}{\textbf{Accuracy}} & \multicolumn{3}{c}{\textbf{Overall}} \\
        \cmidrule(lr){4-6} \cmidrule(lr){7-9}
        & & & Precision & Recall & $\text{F}_3$ & $SR\uparrow$ & $RC\uparrow$ & $DS\uparrow$  \\
        \midrule
        100\% & 100\% & 100\% & 0.667 & 0.667 & 0.667 & 40.00 & 94.33 & 77.48 \\
        160\% & 300\% & 200\% & 0.429 & 1.000 & 0.882 & 20.00 & 83.72 & 63.93 \\
        \rowcolor[gray]{0.9} 130\% & 200\% & 150\% & 0.667 & 1.000 & 0.952 & 60.00 & 98.67 & 83.61  \\
        \bottomrule
      \end{tabular}
      \end{adjustbox}
  }
  \vspace{-10pt}
\end{table}

\mr{Table~\ref{rq4:table0} presents the results of our parameter sensitivity analysis using \UniADtools. Results of other ADSs are provided at our replication website~\cite{website1} due to space limitation.
For buffer settings, smaller buffers yield consistently higher recall but lower precision, limiting the overall DS metric. Our adopted setting (\ie $M$=5, $l$=4, and $R$=20) achieves a balanced trade-off with the best overall performance. For bounding box enlargement factors, moderate enlargement ratios outperform both overly conservative and aggressive settings.}

\mr{\textit{\textbf{Summary.}} \tool maintains improved resilience across different parameter settings, though overly aggressive or conservative settings cause false positives or missed hazards.}


\subsection{\mr{Generalization Evaluation (RQ6)}}

\begin{table}[!t]
    \caption{\mr{Results of the Generalization Evaluation}}
    \vspace{-5pt}
    \label{rq6:table1}
    \centering
    \begin{adjustbox}{width=0.95\linewidth}
    \begin{tabular}{lcccccc}
        \toprule
        \multirow{2}{*}{\textbf{Approach}} & \multicolumn{3}{c}{\textbf{Bench2Drive220}} & \multicolumn{3}{c}{\textbf{CLV20}} \\
        \cmidrule(lr){2-4} \cmidrule(lr){5-7}
         & $SR\uparrow$ & $RC\uparrow$ & $DS\uparrow$ & $SR\uparrow$ & $RC\uparrow$ & $DS\uparrow$ \\
        \midrule
        \textbf{Apollo} & 12.73 & 47.61 & 37.16 & 0.00 & 2.12 & 0.74\\
        \textbf{Apollo-REDriver} & 30.45 & 66.32 & 52.44 & 0.00 & 2.97 & 1.53 \\
        \rowcolor[gray]{0.9} \textbf{\Apollotools} & 36.81 & 71.30 & 60.56 & 0.00  & 4.52 & 1.96 \\
        \bottomrule
    \end{tabular}
\end{adjustbox}
\end{table}

\mr{Table~\ref{rq6:table1} presents the overall effectiveness of \tool when generalized to the modular ADS, \ie Apollo. Compared to the original version of Apollo, \Apollotools improves the $SR$, $RC$, and $DS$, on average, by 94.58\%, 81.48\%, and 113.92\% on these two benchmarks, respectively, achieving additional gains of 10.44\%, 19.85\%, and 21.79\% over Apollo-REDriver.}

\mr{\textit{\textbf{Summary.}} \tool shows a strong generalization capability to modular ADSs, consistently outperforming both the original Apollo and Apollo-REDriver across different benchmarks.}


\section{Threats to Validity}
First, the quality of BEV representations poses a potential threat to validity. To mitigate this threat, we adopt BEV representations generated by BEVFormer to demonstrate the practical feasibility of our approach under realistic perception conditions. In addition, we leverage privileged BEV representations derived from simulator-provided ground-truth information to validate the upper-bound effectiveness of our framework. We believe that as BEV perception techniques continue to improve, the real-world applicability of \tool will be further~enhanced.

Second, the diversity of driving hazards presents another threat. Our current implementation focuses on three critical hazards, \ie the collision hazard, the stop signal hazard, and the stalling hazard.
Our modular design and extensible architecture allow \tool to be expanded to support additional hazards, such as making illegal U-turns in no-U-turn zones, or entering time-restricted entry zones outside permitted hours, where access is controlled based on a predefined schedule (\eg school zones or peak-hour bus lanes).
While these categories cover a wide range of common safety violations, other hazards, such as sensor faults or control delays, lie beyond the scope of our current design and would require separate modeling approaches.

Third, the selection of benchmarks and target ADSs introduces a potential threat to external validity.
\mr{To mitigate this threat, we evaluate \tool on two widely used benchmarks and three representative end-to-end ADSs, capturing the diversity of driving environments and ADS architectures.
We conduct~a paired t-test analysis~\cite{box1987guinness} across the two benchmarks with diverse towns and weather conditions. 
\tool significantly improves resilience with p-values of 4e-10 and 3e-4 (p$<$0.05) over various towns and weather conditions, respectively.}
Besides, \tool and \tools achieve average p-values of 0.026 and 0.020 on resilience metrics over all target ADSs, respectively, indicating statistically significant improvements~(p$<$0.05). 
These results demonstrate the robustness and general applicability of \tool across diverse ADSs and driving environments.
Therefore, we believe that \tool can be effectively adapted to other end-to-end ADSs to achieve similar improvements.
Moreover, since the driving hazards addressed in our work are common in real-world traffic, we believe that \tool has strong potential~to~enhance~resilience~in~real-world~ADSs.

\mr{Finally, the selection of metrics introduces a potential threat to measuring software resilience. To mitigate this threat, we define software resilience as the ability to adapt to and recover from unexpected events while maintaining effective operation under hazardous conditions. Accordingly, we select a broad set of evaluation metrics. Specifically, we assess resilience in terms of hazard adaptation (\ie success rate, route completion, driving score, violation prevention, and takeover accuracy) and recovery efficiency (\ie negligible switching latency).} 


\section{Related Work}

\subsection{Resilience in Autonomous Systems}
\mr{Resilience in autonomous systems~\cite{bagchi2020vision,matthews2016resilient, desai2019soter} has been widely studied, aiming to enable systems to adapt to and recover~from various hazards.}
Yang et al.~\cite{yang2021enabling} enhance the fault tolerance of general-purpose graphics processing unit by selectively replicating thread warps vulnerable to transient faults. MAVFI~\cite{hsiao2023mavfi} employs a two-layer autoencoder to monitor the velocity and the time to collision of unmanned aerial vehicles, and triggers the signal to halt the propagation of~errors. 

In the domain of ADSs, Xia et al.~\cite{xia2025robust} propose a long-range perception system that is robust against sensor misalignment. Meanwhile, Waymo~\cite{waymo2017waymo} enhances system resilience by deploying redundant hardware that supports immediate failover to maintain safe operation during faults. However,~a~recent work~\cite{wan2022analyzing} highlights that evaluating resilience in isolated components of the compute or control stack lacks a holistic, cross-stack perspective, potentially overlooking error propagation and limiting the effectiveness of resilience solutions.
To the best of our knowledge, there is no existing work focusing on enhancing the resilience of end-to-end ADSs from a system-level perspective.
In this work, we enhance the resilience of end-to-end ADSs at a system level by enabling continuous hazard monitoring and proactive mitigation.

\subsection{Safety Assurance in Autonomous Driving Systems}
Ensuring that ADSs operate within strict safety bounds is essential for safeguarding both passengers and other road users, particularly in hazardous scenarios~\cite{gyllenhammar2025road,gyllenhammar2021minimal,watanabe2018runtime}.\mr{~To~monitor driving hazards, prior studies~\cite{hussain2022deepguard, stocco2022thirdeye, stocco2020misbehaviour, shao2024likely} have explored predicting unexpected driving conditions based on distributional shifts or low model confidence.} However, such approaches are often impractical for real-time, resource-constrained environments~\cite{grewal2024predicting}. 
Alternatively, some works adopt rule-based misbehavior detection~\cite{candela2023risk, candela2021fast, qian2025collision, yu2024online, quinonez2020savior}, which leverage vehicle dynamics to identify potential hazards before safety violations occur. 
However, such vehicle dynamics-based approaches often depend on accurate physical parameters (e.g., road friction, vehicle mass) and require more computation than the KBM we adopt.
Besides, these works primarily focus on detecting hazards and do not provide solutions for adaptive responses to ensure safety. In contrast, our work presents a runtime framework incorporating both continuous hazard monitoring and adaptive hazard mitigation to ensure safety.

Beyond hazard monitoring, some works~\cite{sun2024redriver,sun2025fixdrive} directly intervene the planning process by modifying each trajectory~to enforce specification compliance under the assumption of perfect perception. However, these approaches are tightly~coupled with a specific system named Apollo, a modular~pipeline ADS, and are difficult to  generalize to end-to-end ADSs. Differently, our work is a general approach for end-to-end ADSs.
Furthermore, various fallback strategies have been proposed to ensure driving safety under hazardous conditions~\cite{yu2019fallback, hussain2022deepguard, xue2018fallback,grieser2020assuring,cheng2020guardauto}. 
\mr{Simplex-Drive~\cite{chen2022runtime} and RTA-IR~\cite{peng2023rta} focus on avoiding vehicle collisions under \textit{highway-env}~\cite{highway-env}, without considering the~real perception of the sensor.}
Dual-AEB~\cite{zhang2024dual} leverages large language models to trigger appropriate braking responses, but ultimately relies on human-initiated takeovers~\cite{gold2013take, yu2022remote}. Xue et al.~\cite{xue2018fallback} further explore fallback strategies that aim to park the vehicle in safe areas during emergencies. 
However, these approaches primarily emphasize emergency intervention without addressing continuous safety assurance through hazardous scenarios.
In contrast, we propose a framework to enhance the resilience of end-to-end ADSs, particularly the ability to continuously monitor hazards, adaptively respond to potential safety violations, and recover quickly to sustain safe operation in hazardous scenarios.

\section{Conclusion}
We have proposed and implemented a resilience-oriented runtime framework, \tool, to mitigate the driving hazards, thus
preventing potential safety violations and improving the driving performance of end-to-end ADSs. 
Large-scale experiments have been conducted to demonstrate the effectiveness and efficiency of \tool.


\section*{Acknowledgment}
This work was supported by the National Natural Science Foundation of China (Grant No. 62332005 and 62372114).

\clearpage

{\footnotesize
\bibliographystyle{IEEEtranS}
\bibliography{IEEEabrv,src/reference}
}

\end{document}